\documentclass{article}
\usepackage{acra}

\usepackage{subcaption}
\usepackage{booktabs}
\usepackage{balance}
\usepackage{color}
\usepackage{tabularx}
\usepackage{graphicx,dblfloatfix} 
\usepackage{amsmath}
\usepackage{amssymb}
\usepackage{listings}
\usepackage{textcomp}
\usepackage{url}
\usepackage{multirow}
\usepackage{todonotes}
\usepackage{wrapfig}
\usepackage{ragged2e} 
\usepackage[subnum]{cases}
\newcolumntype{Y}{>{\RaggedRight\arraybackslash}X} 
\usepackage{algorithm}
\usepackage[noend]{algpseudocode}
\usepackage{tikz}
\usepackage{tkz-euclide}
\usetikzlibrary{math}
\usepackage{pgfplots}
\usetikzlibrary{arrows}
\usepackage{float}

\newcommand{\ts}{\textsuperscript}

\title{Autonomous Obstacle Legipulation with a Hexapod Robot}
\author{Bethany Lu$^{1,2}$ , Benjamin Tam$^{1,3}$, Navinda Kottege$^{1}$ \\ $^{1}$Robotics and Autonomous Systems Group, CSIRO, Pullenvale, QLD 4069, Australia \\ 
$^{2}$University of Technology Sydney, Ultimo, NSW 2007, Australia \\ $^{3}$School of Information Technology and Electrical Engineering, \\ University of Queensland, St Lucia, QLD 4072, Australia}

\begin{document}

\maketitle

\begin{abstract}
Legged robots traversing in confined environments could find their only path is blocked by obstacles. In circumstances where the obstacles are movable, a multilegged robot can manipulate the obstacles using its legs to allow it to continue on its path. We present a method for a hexapod robot to autonomously generate manipulation trajectories for detected obstacles. Using a RGB-D sensor as input, the obstacle is extracted from the environment and filtered to provide key contact points for the manipulation algorithm to calculate a trajectory to move the obstacle out of the path. Experiments on a 30 degree of freedom hexapod robot show the effectiveness of the algorithm in manipulating a range of obstacles in a 3D environment using its front legs.
\end{abstract}

\section{Introduction}
\label{sec:introduction}

Terrains in disaster zones, subterranean environments and vegetated areas present a combination of fixed obstacles, movable obstacles and irregular ground. Robots could be prohibited from navigating confined spaces when the only path is blocked by movable obstacles. Wheeled and tracked robots have limited ability to traverse these challenging environments as they require continuous ground contact points which can damage the terrain. On the other hand, legged robots can place their foot tips on small footholds in discontinuous terrain \cite{tennakoon_safe_2020b}, adjust their footprint to pass through confined areas \cite{buchanan_walking_2019}, and traverse rough terrain \cite{bjelonic_weaver:_2018}. To successfully traverse these unstructured terrains with unknown obstacles in the robot's path, the robot is required to manipulate obstacles out of its way. Thus, if a legged robot platform is able to autonomously identify and manipulate an obstacle in its path, the robot can progress further in the environment. 

\begin{figure}[t!]
\centering
    \includegraphics[width=1\linewidth]{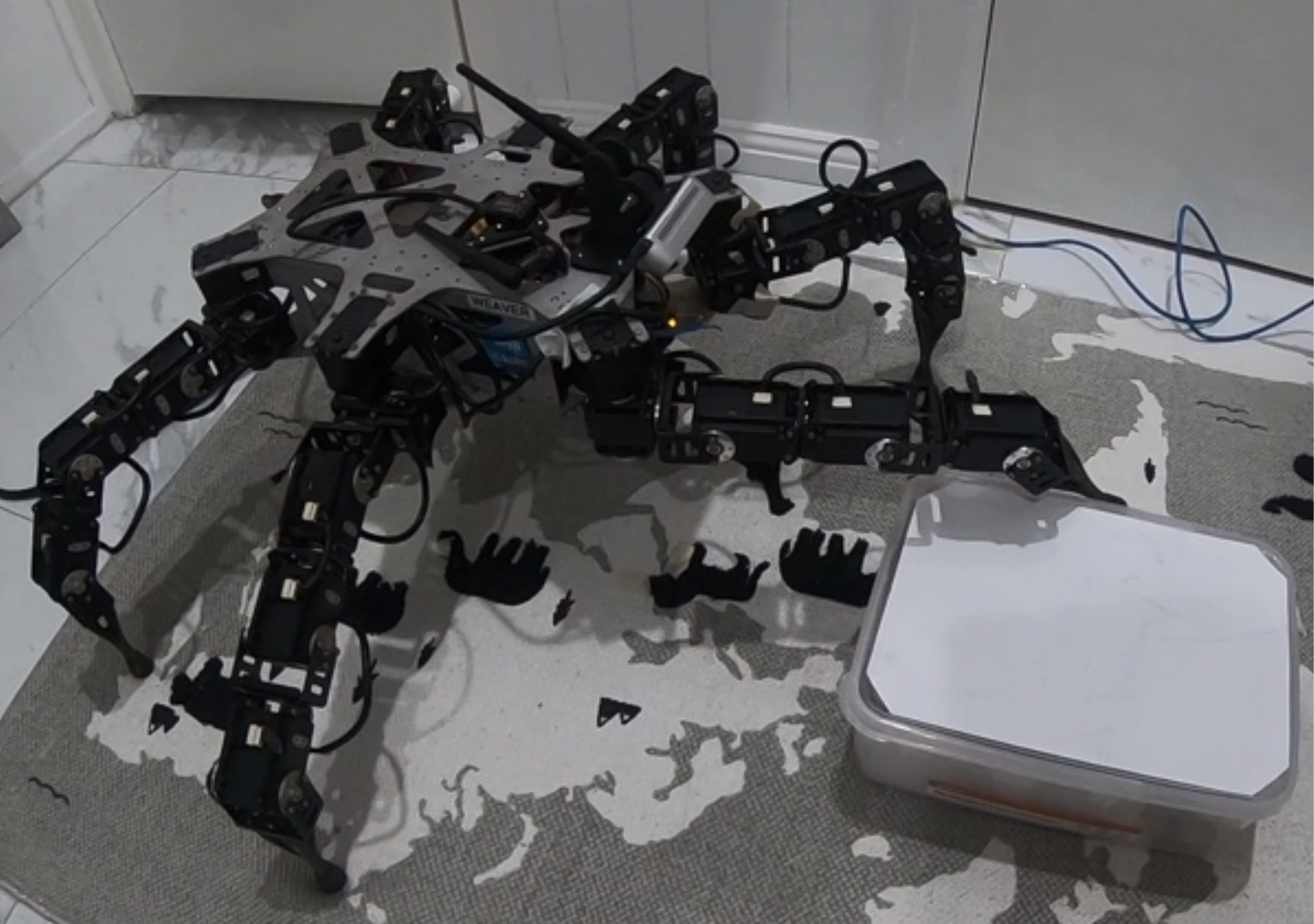}
    \caption{Hexapod robot Weaver manipulating an obstacle.}
    \label{fig:heroweaver}
\end{figure}

Existing legged mobile manipulators have attached gripper equipped manipulator arms onto agile quadruped platforms \cite{bellicoso_alma_2019,boston_dynamics_spot_2019,rehman2016towards}. 
This method provides dexterous manipulation capability at the expense of reduced payload and decreased operating time due to the additional mass. While some dynamic scenarios would benefit from a fully integrated manipulator, these disadvantages have lead researchers to focus on utilising the legs for the dual purpose of mobility and manipulation, or legipulation. Quadruped robots are only able to manipulate with a single leg while standing stationary, or with two legs while sitting \cite{hebert_mobile_2015}. On the other hand, hexapod robots, such as Weaver shown in Figure~\ref{fig:heroweaver}, have the advantage of grasping objects with up to 2 legs and still being able to walk with statically stable gaits. Hexapod robot platforms such as LAURON V \cite{roennau_lauron_2014}, LEMUR-II \cite{kennedy_lemur_2006}, MAX \cite{Elfes2017}, MELMANTIS \cite{koyachi_control_2002}, and ASTERISK \cite{Takubo_2006jrm} have demonstrated manipulating objects with legs. 

LAURON V used a RGB-D camera system to detect objects of interest \cite{heppner_laurope_2015} and predefined grasp trajectories for a gripper \cite{HEPPNER_versatile_2014} attached to the front right leg to pick and store objects. A stereo camera pair on LEMUR-II was used to detect fiducial markers attached to the objects and arm to reduce pose errors \cite{nickels_vision-guided_2006}. The vision algorithms performed visual servoing to achieve autonomous docking and bolt fastening.  
To grasp and move objects simultaneously, various gripper designs \cite{lewinger_insect-inspired_2006} and locomotion gaits \cite{deng_object_2018} have been investigated. To move large objects, \cite{inoue_pushing_2010} developed a novel approach of utilising two upper legs and the robot's body to increase exertion force.
Two legged manipulation through a combination of teleoperation and predefined motions have been shown in \cite{koyachi_control_2002,Takubo_2006jrm}.

Our approach is similar to \cite{heppner_laurope_2015} where we use a RGB-D sensor to detect the object pose and use inverse kinematics to move the leg to the object, without the use of fiducial markers and visual servoing \cite{nickels_vision-guided_2006} for pose tracking. While the previous works have focused on either predefined motions for known objects or teleoperation of unknown objects, our approach extends robot capability through the calculation of control points for the trajectory based on the object size and location from point cloud data. This allows for autonomous manipulation for different sized obstacles using different legs without a gripper.

\begin{figure}[t!]
\centering
    \includegraphics[width=60mm]{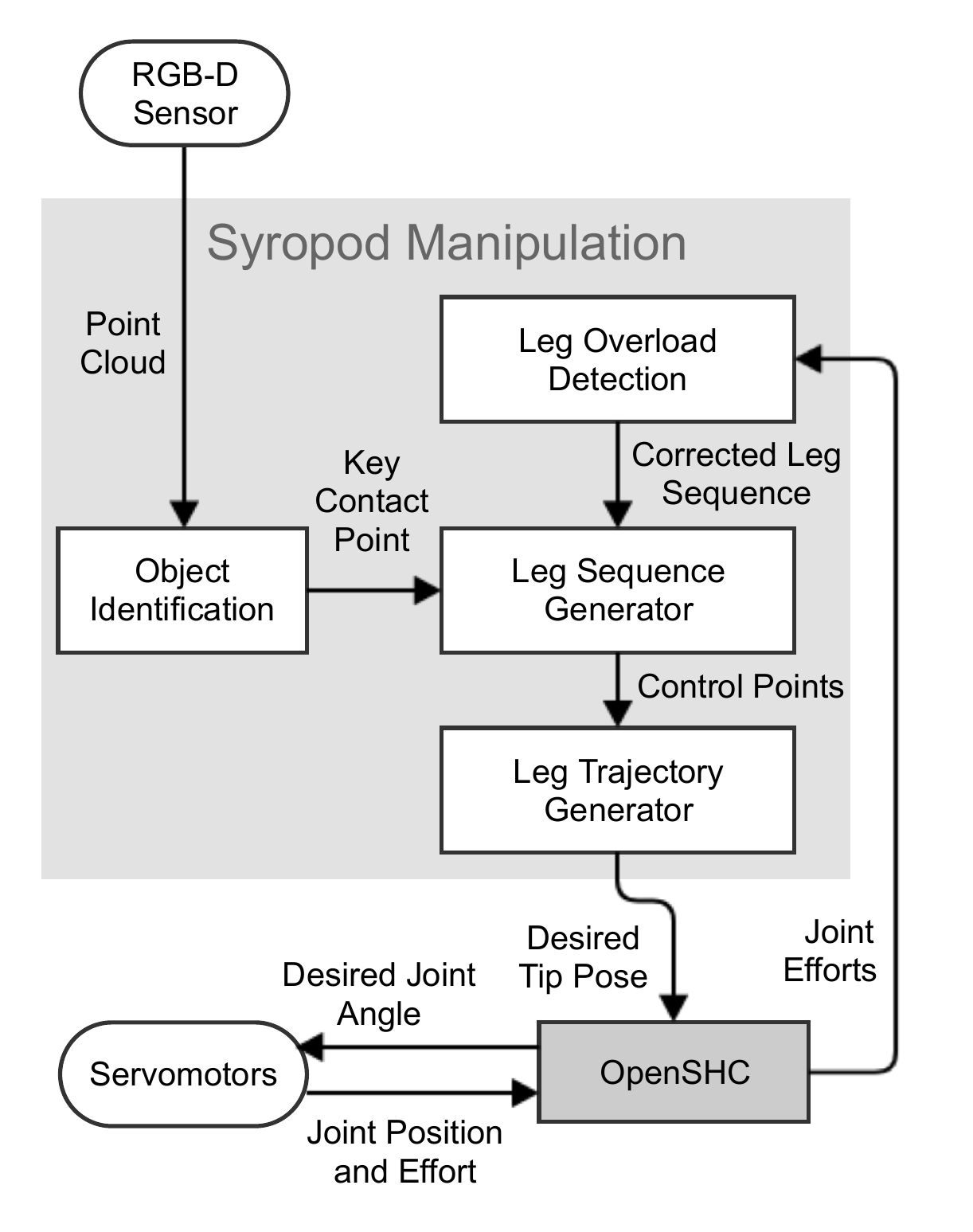}
    \caption{Overview of the Syropod Manipulation system.}
    \label{fig:system_overview}
\end{figure}

In this paper, we present Syropod Manipulation, a framework integrating perception and manipulation on a hexapod robot to achieve autonomous legipulation of obstacles in the robot's path. A RGB-D sensor is used to detect obstacles in the robot's path, with the leg trajectory generated to sweep obstacles away. The framework focuses on obstacles that can be moved away from the region in front of the robot with a single pushing motion.

We detail the framework in Section \ref{sec:approach} for the perception of obstacles and the generation of trajectories for legipulation. Section \ref{sec:experiments} details the experiments conducted, with results shown in Section \ref{sec:results} and discussed in Section \ref{sec:discussion}. The paper is concluded in Section \ref{sec:conclusions}.


\section{Syropod Manipulation}
\label{sec:approach}

Syropod Manipulation is comprised of Obstacle Identification and Legipulation, which are the perception and manipulation modules respectively, as shown in Figure~\ref{fig:system_overview}. Obstacle Identification utilises a RGB-D camera to gather point cloud data of the environment and extract the obstacle location. The processed data is passed into Legipulation to control how the leg will interact with the obstacles. A high-level controller such as OpenSHC \cite{tam2020openshc} controls the robot servomotors to achieve the desired tip pose.

\subsection{Obstacle Identification}
\label{sec:obstacleID}

\begin{figure}[t!]
\centering
    \includegraphics[width=85mm]{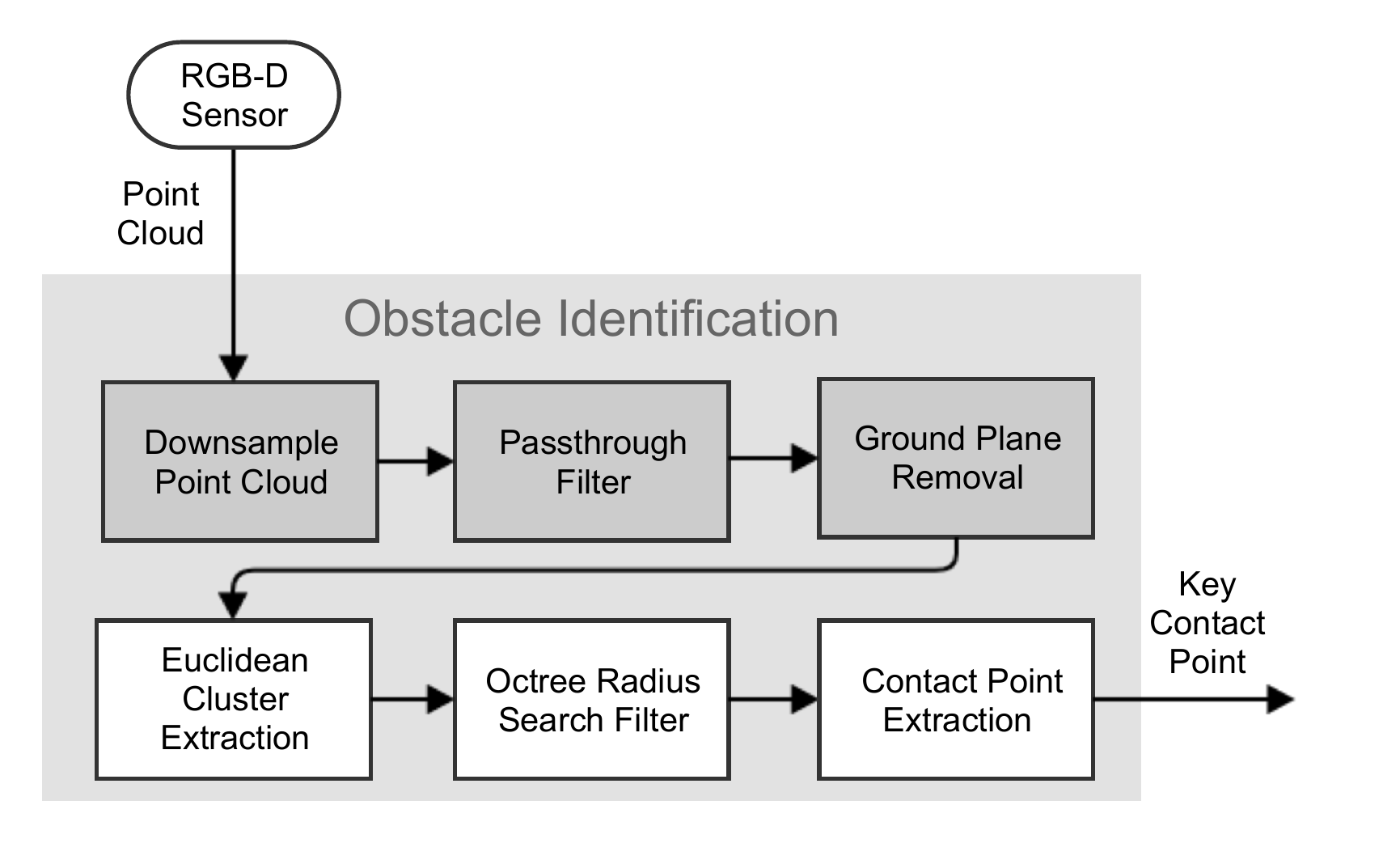}
    \caption{Overview of the Obstacle Identification perception system.}
    \label{fig:perpception_overview}
\end{figure}

The perception system uses a RGB-D sensor to identify and isolate obstacles where manipulation is feasible based on its relative size to the robot from the surrounding scene. Obstacle identification uses point cloud data to extract the obstacle directly in front of the robot from the environment. Our work is based on \cite{Zeineldin2016} with several modifications, as outlined in Figure~\ref{fig:perpception_overview}, to extract the key contact point for manipulation. For obstacles 0.2\,m to 0.4\,m away from the sensor, the obstacles occupy the majority of the sensor's field of view. Thus, the point cloud is downsampled to 0.01\,m voxels to reduce computation without affecting accuracy. A passthrough filter is used to remove points outside the workspace of the front legs of the robot. Then, the ground plane is removed via RANSAC to leave the remaining points that represent the obstacles. These modules follow \cite{Zeineldin2016} and appear shaded in Figure~\ref{fig:perpception_overview}.

We extend the work in \cite{Zeineldin2016} to isolate the closest obstacle from all obstacles detected. This allows the robot to sequentially manipulate each obstacle in its path. Additional filtering with Euclidean Cluster Extraction and Octree Radius Search is utilised to detect at close proximity the location of the target obstacle. A bounding box is fitted to the obstacle's point cloud and the key contact point for the robot to manipulate is calculated in the Contact Point Extraction module.

\subsubsection{Euclidean Cluster Extraction}
\label{sec:euclidean_cluster}
The point cloud without the ground plane is clustered into groups which identifies different focus areas within the field of view.  This filter groups neighbouring clusters within a threshold together. This further removes any outliers that do not belong to the group of clusters. The robot only needs to manipulate obstacles which are large enough to cause potential issues when traversing, as smaller obstacles can be stepped over. Thus, the parameters for the Euclidean Cluster Extraction from the Point Cloud Library \cite{Rusu_ICRA2011_PCL} were empirically selected based on observations with the robot.

\subsubsection{Octree Radius Search Filter}
\label{sec:octree_filter}
\begin{figure}[t!]
\centering
    \includegraphics[height=50mm]{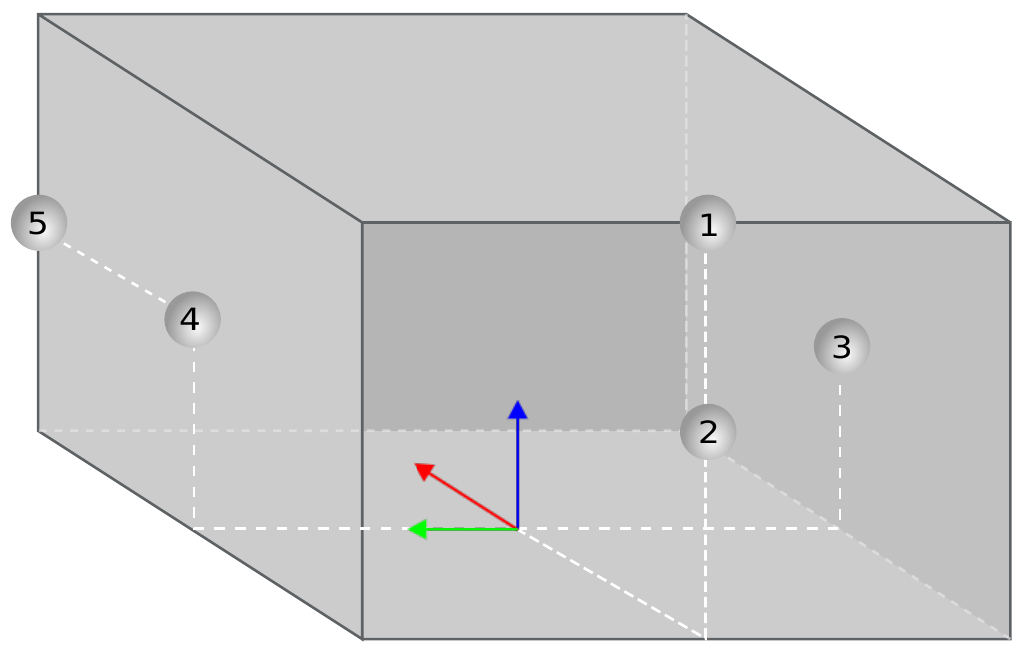}
    \caption{Key object contact points based on the bounding box. Point 1: Front top centre; Point 2: Front centre; Point 3: Right centre; Point 4: Left centre; Point 5: Left back centre.}
    \label{fig:bounding_box}
\end{figure}
The clusters are filtered for the nearest neighbours at the target search location within a specified radius.

The filter is initially given a $(x, y, z)$ coordinate in front of the robot at the centre to search for objects. The search location is incrementally increased from the centre to the peripherals of the field of view until the closest obstacle is found.

These additional filters removes stray clusters that can cause the bounding box for the object to be inflated. Thus, the euclidean cluster extraction arranges the major clusters into groups, with the octree filter searching within these groups to single out the object.

\subsubsection{Contact Point Extraction}
\label{sec:feature_extraction}

\begin{figure}[t!]
\centering
    \includegraphics[height=50mm]{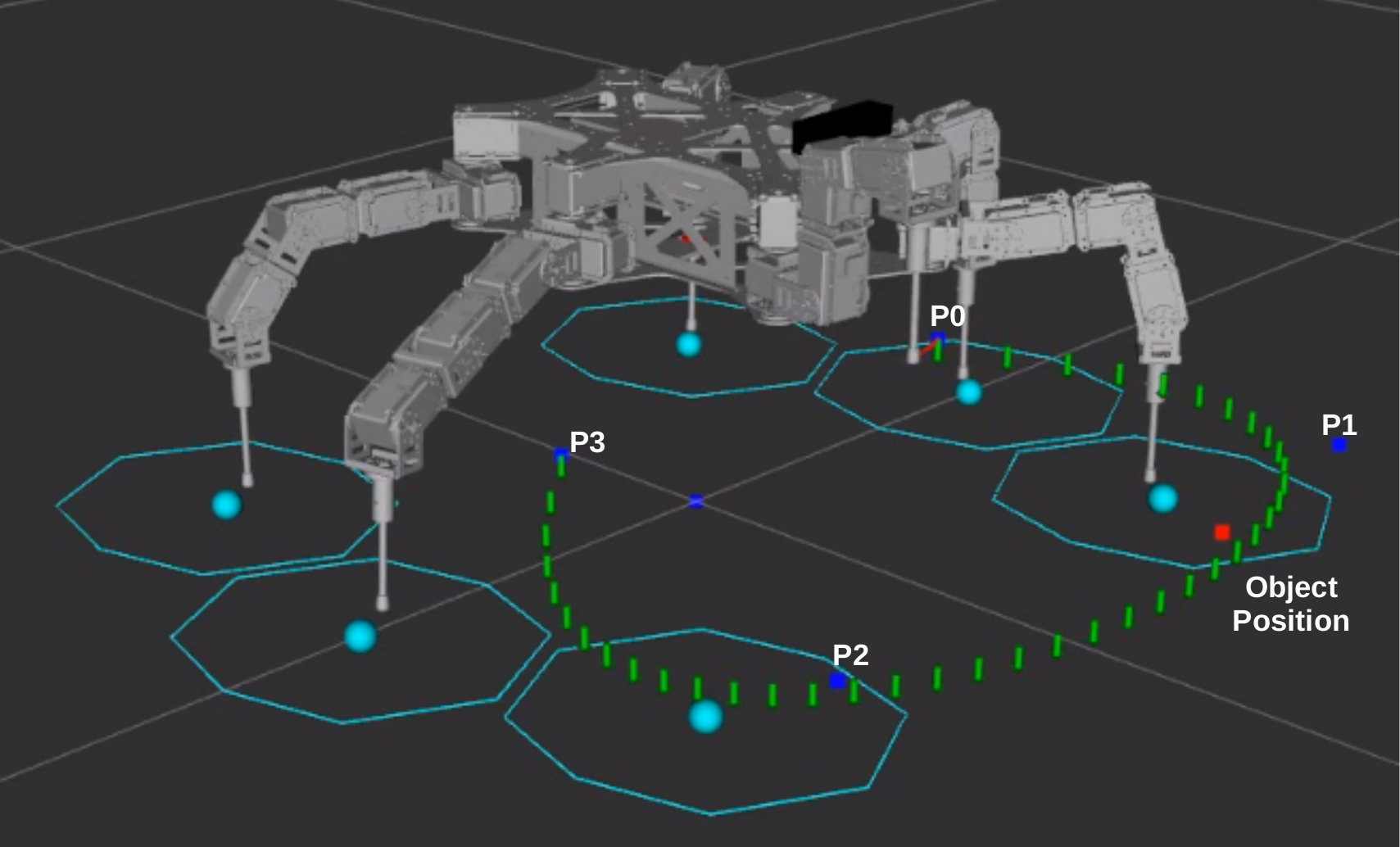}
    \caption{Weaver simulation following the trajectory generated by the modified cubic B\'{e}zier equation. Downward arrows indicate the location and orientation for the leg tip to follow.}
    \label{fig:weaver_control_point_test_sim}
\end{figure}
The identified closest object is surrounded by a bounding box. The bounding box extracts the height and width of the detected object and provides key points where the robot can interact with the object, as shown in Figure~\ref{fig:bounding_box}. The key point on the boundary box is selected based on the legipulation behaviour and is mapped to the object.
The selected key point for each respective legipulation behaviour is predefined. The key points provides the position of the object and the path for the robot’s leg to pass through. 

The location of the object influences the final position of the intermediate and last control points for trajectory generation, highlighted by the red dot influencing the position of $P_1$, $P_2$ and $P_3$ in Figure~\ref{fig:weaver_control_point_test_sim}. 
Additional key contact points can be specified for complex interaction motions such as combining lifting and pushing. The key contact points are defined prior to executing manipulation.

\subsection{Legipulation}
\label{sec:legipulation}

Spatial control of the leg allows unique leg movements for interacting with different objects. A single or a combination of splines are used to create the trajectory which guides the leg tip to the desired locations. Splines formed by B\'{e}zier curves are used to generate the desired smooth trajectory and Spherical linear interpolation (Slerp) is used to define the desired final orientation of the leg tip. 
The key contact point from the object is fed into the control points of the curves, guiding the leg to interact with the object. 

\subsubsection{Leg Trajectory Generation}
\label{sec:leg_trajectory_generation}
B\'{e}zier curves are used to control the position of the leg trajectory in 3D space while Slerp is used to interpolate from the current to desired leg tip rotation. The combination of both allows the pose of the leg tip to be defined. Thus, the leg tip can remain at a particular orientation throughout the execution if required and the degrees of freedom allow. B\'{e}zier curves generate smooth transitions for the tip position  from the start to final position. The equations defining quadratic (2\ts{nd} order) and cubic (3\ts{rd} order) B\'{e}zier curves are given by:

\begin{equation}
    ^2B(t) = s^2P_0 + 2tsP_1+t^2P_2
    \label{eq:2ndbeziercurve}
\end{equation}

\begin{equation}
    ^3B(t) = s^3P_0 + 3ts^2P_1+3st^2P_2+t^3P_3
    \label{eq:3rdbeziercurve}
\end{equation}

respectively, where $s = 1-t$ and $t \in [0,1]$. $P_0$ is control point 0 (starting point); $P_1$ is control point 1; $P_2$ is control point 2; and $P_n$ is control point $n$ (final point) for $^nB(t)$.

\begin{figure}[t!]
     \centering
         \begin{subfigure}[b]{0.5\linewidth}
         \includegraphics[width=\linewidth]{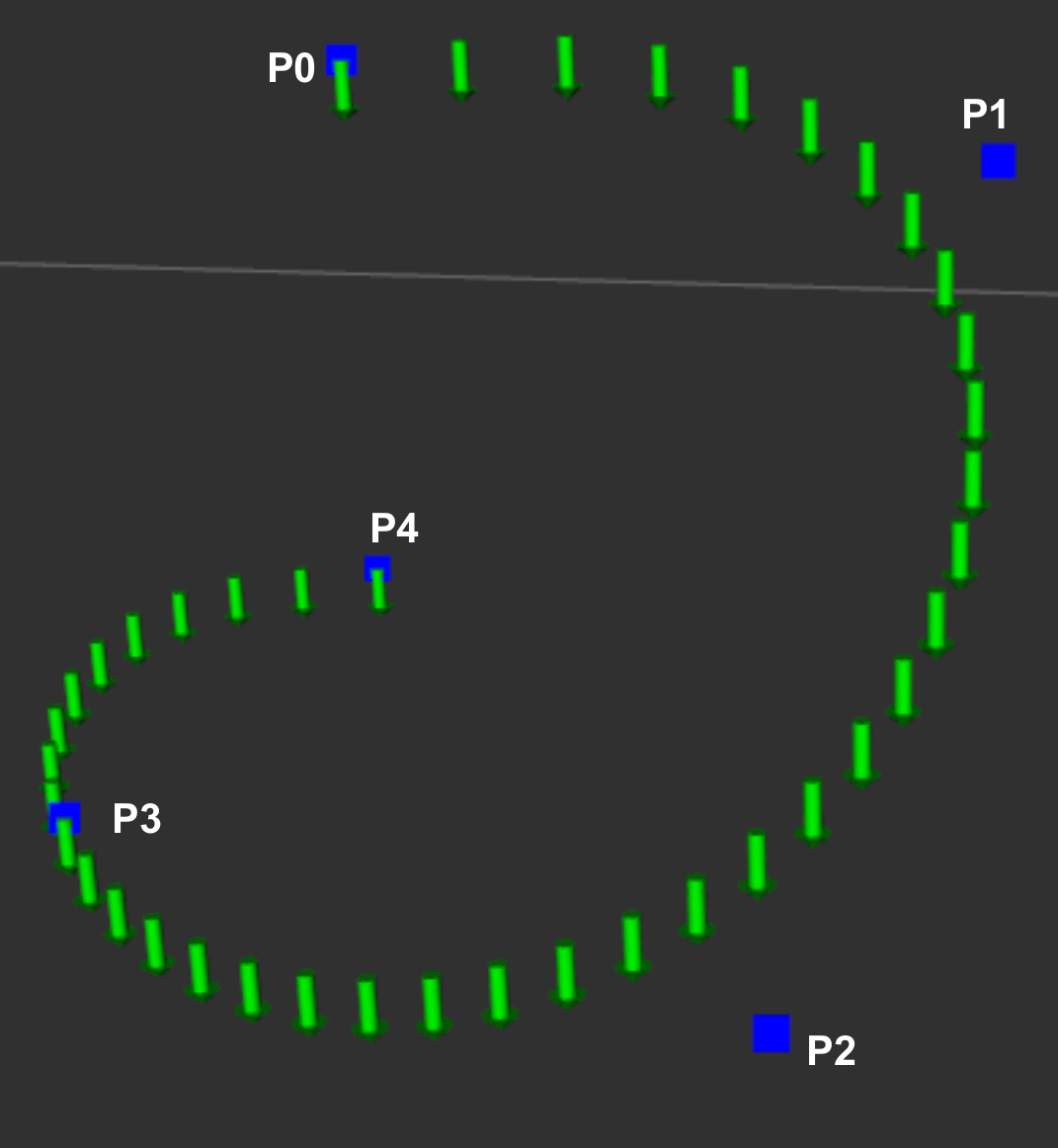}
         \caption{}
         \label{fig:high_vs_low_order_curve_2}
     \end{subfigure}
     \begin{subfigure}[b]{0.458\linewidth}

         \includegraphics[width=\linewidth]{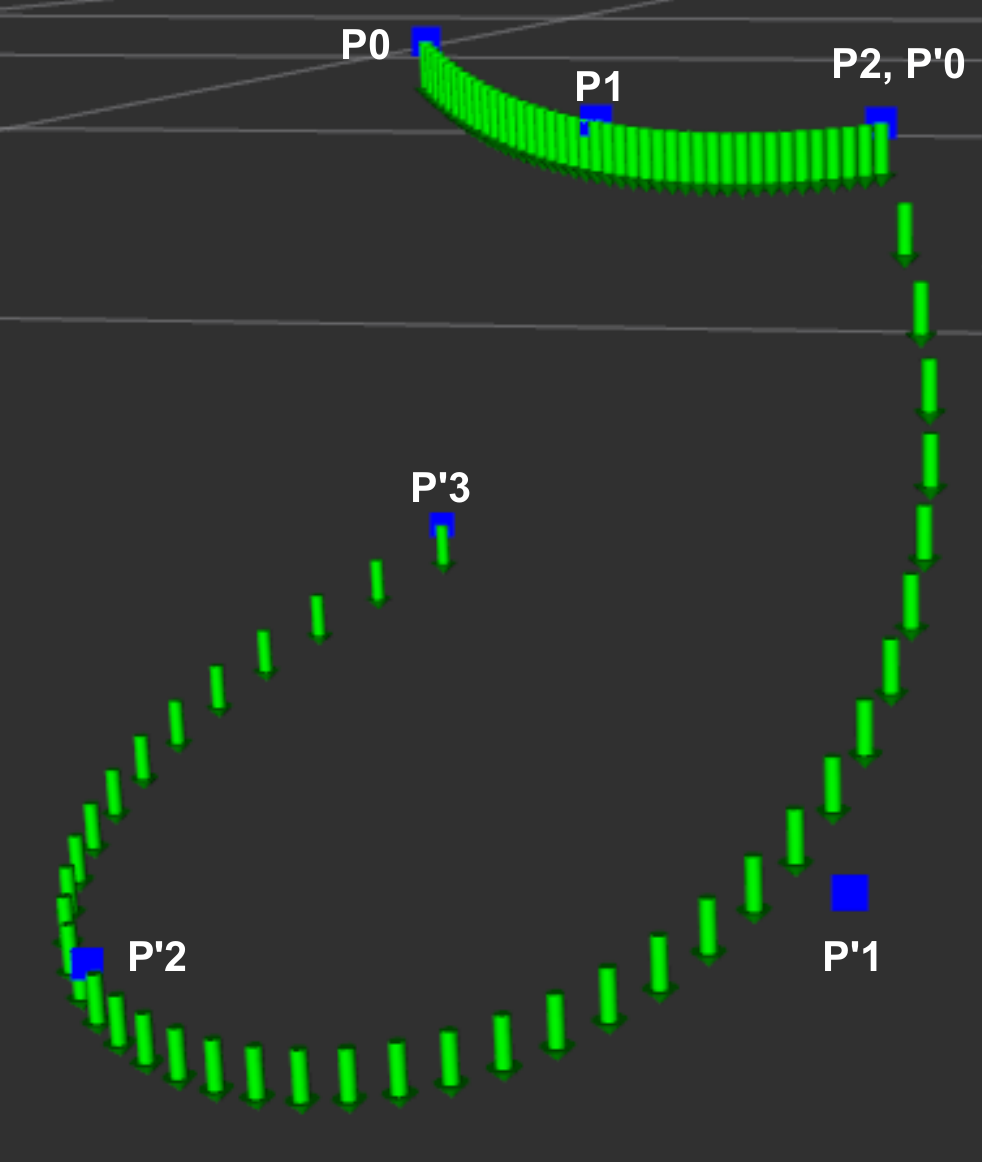}
         \caption{}
         \label{fig:high_vs_low_order_curve_1}
     \end{subfigure}
        \caption{(a) A Quartic B\'{e}zier curve and (b) a sequence of Quadratic \& Cubic B\'{e}zier curves.}
        \label{fig:high_vs_low_order_curve}
\end{figure}

The control points defining the B\'{e}zier curve do not normally lie on the curve itself but rather a certain distance away.
Thus, not all the control points can be used as desired points, locations where we want the curve to pass through.

For all control points to be used as desired points, the B\'{e}zier curve Equations \ref{eq:2ndbeziercurve} and \ref{eq:3rdbeziercurve} are modified to recalculate $P_1$ and $P_2$.

The modified point $\hat{P_1}$ for the quadratic B\'{e}zier curve from Equation \ref{eq:2ndbeziercurve} is given by:
\begin{equation}
    \hat{P_1} = \frac{^2B(t) - s^2P_0 - t^2P_2}{2ts}
    \label{eq:2ndmodbeziercurve}
\end{equation}
The modified points $\hat{P_1}$ and $\hat{P_2}$ for the cubic B\'{e}zier curve from Equation \ref{eq:3rdbeziercurve} is given by:
\begin{equation}
    \hat{P_1} = \frac{^3B(t) - s^3P_0 -3st^2P_2-t^3P_3}{3ts^2}
    \label{eq:3rdmodbeziercurvep1}
\end{equation}
\begin{equation}
     \hat{P_2} = \frac{^3B(t) - s^3P_0 - 3ts^2\hat{P_1}-t^3P_3}{3st^2}
    \label{eq:3rdmodbeziercurvep2}
\end{equation}

For Equations \ref{eq:2ndmodbeziercurve} and \ref{eq:3rdmodbeziercurvep1}, $\hat{P_1}$ takes in the original control points. For Equation \ref{eq:3rdmodbeziercurvep2}, $\hat{P_2}$ takes in the modified control point $\hat{P_1}$ and the remaining original points. Thus, the changes for $\hat{P_1}$ also affects $\hat{P_2}$. These modification allows us to obtain the results in Figure~\ref{fig:high_vs_low_order_curve}, where the curve is approximately nearer or on the desire points.

\begin{table*}[b]
\centering
\caption{Dimensions follow L x W x H unless otherwise specified. Location of the object in $(x,y)$ coordinate relative to the centre of the robot body.}

\begin{tabular} {c c c c c}

\toprule
Object & Weight (kg) & Dimension (mm) & Location (mm) & Surface\\
\midrule
Object 1 & 0.013  & Dia. 66.2 x 115.2 & (360, -30) & Carpet \\
Object 2 & 0.39   & 300 x 224 x 115 & (360, -30) & Carpet, Concrete, Marble\\
Object 3 & 1.45 & 300 x 224 x 115 & (360, -30) & Carpet\\
\bottomrule
\end{tabular}
\label{tab:object_details}
\end{table*}

\subsubsection{Leg Sequence Generation}
\label{sec:leg_sequence}
Each control point on the B\'{e}zier curve defines a future position the leg tip will visit. A leg trajectory is generated with control points defined for:
\begin{enumerate}
\item Initial position - The current pose of the selected leg tip.
\item Safe position - A predefined and tested pose where it will not damage the robot or object.
\item Beside the object - A pose where the leg is ready to interact with the object/obstacle.
\item Final position - A pose which completes the entire motion or the final pose of the leg tip.
\end{enumerate}
The order of the B\'{e}zier curve used can be modified. Depending on the type of motion, it is beneficial to use sequenced cubic B\'{e}zier curves to generate the trajectory, rather than a quartic B\'{e}zier curve. This is especially true for simple up and down motion.
Through the use of control points, unique sequences of motion can be created for legged robots. The leg motion can be altered to allow for a modified leg end-effector, such as a gripper.

For a 2\ts{nd} order curve, the generated spline passes through all the control points. However, for complex splines such as 3\ts{rd} and 4\ts{th} order curves, this is not guaranteed, with the spline not passing through all the control points but approximately near it. To generate complex curves while reducing this error, illustrated by $P_0 ... P_4$ in Figure~\ref{fig:high_vs_low_order_curve_2}, the trajectory is segmented into several movements, each defined by a lower order B\'{e}zier curve as illustrated by $P_0 ... P_2$ and $P'_0 ... P'_3$ in Figure~\ref{fig:high_vs_low_order_curve_1}.

\subsubsection{Leg Overload Detection}
\label{sec:leg_feedback}
The system continuously senses whether the obstacle is too heavy to proceed with manipulating. Torque, current or effort feedback from the motors informs the system when the leg is about to be overloaded, allowing the motors to be protected from damage. When overload detection is triggered, the current manipulation action is abandoned and the leg returns to stance position. Additional leg sequences can be executed to attempt moving the obstacle with a different leg configuration instead of returning to stance position. 


\section{Experiments}
\label{sec:experiments}

Syropod Manipulation was deployed on Weaver \cite{bjelonic_weaver:_2018,buchanan_walking_2019,tam_2017}, with an Intel Realsense D435 sensor payload. The algorithm was executed on the onboard Intel i7 PC, powered by an external power supply and connected remotely via an Ethernet cable.
The joint effort and tip positions for each leg were monitored to evaluate the behaviour and response of the leg tip following the leg trajectory motion. 

\begin{figure}[t]
\centering
      \includegraphics[height=55mm]{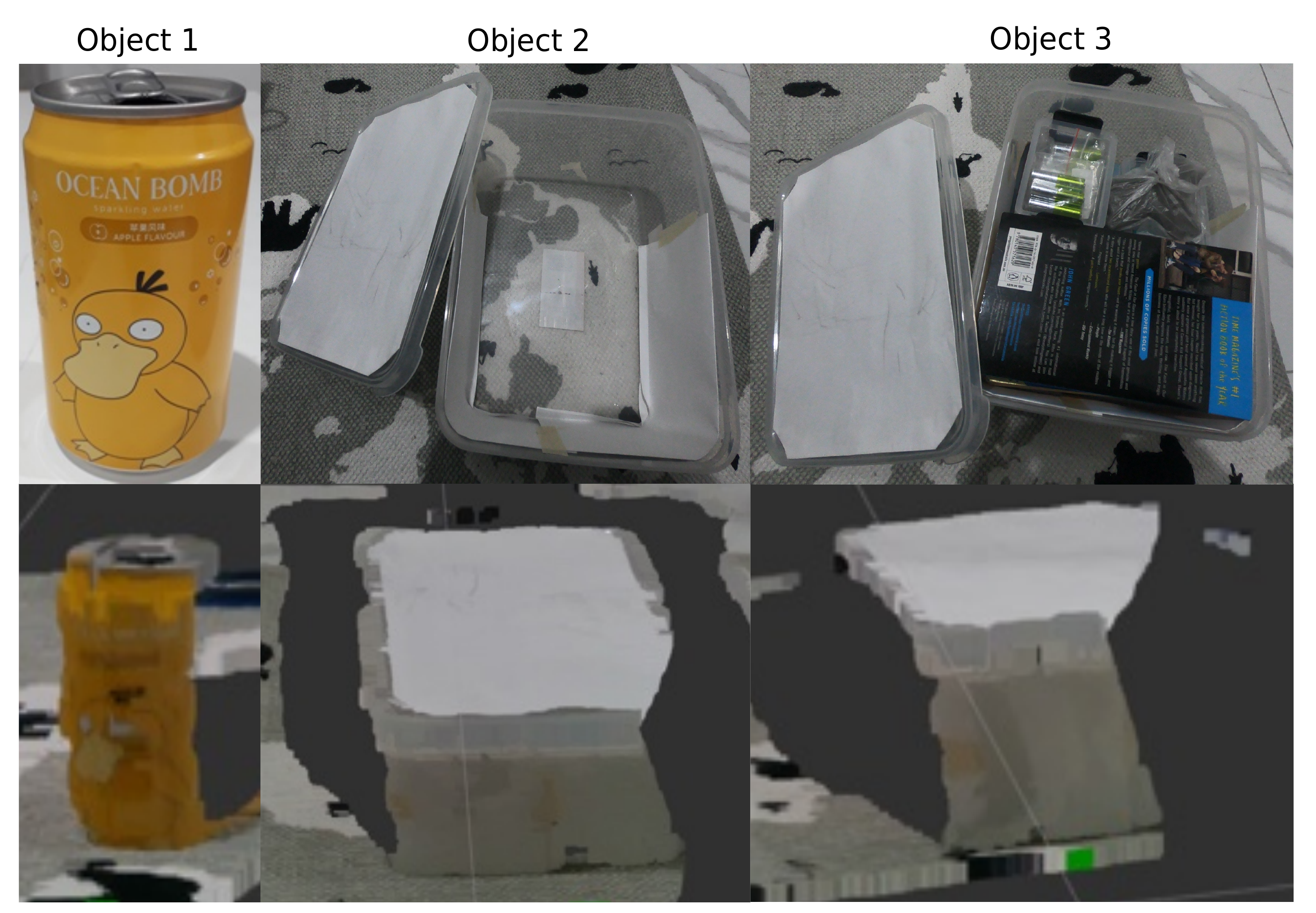}
    \caption{Top row: Images of 3 objects taken by a GoPro camera. Bottom row: Point clouds of the 3 objects seen by the RGB-D camera.}
    \label{fig:all_objects}
\end{figure}

\subsection{Experimental Setup}
\label{sec:experimental_setup}

The objects were selected so that the robot was capable of seeing and interacting with them. Three scenarios tested the effectiveness of the trajectory algorithm as outlined in Table~\ref{tab:object_details}. Specifically, the scenarios are:
\begin{itemize}
\item Object 1 - light weight and small - where the object was detectable and movable. The smaller object size tested accuracy.
\item Object 2 - light weight and large - where the object was detectable and movable. The larger size tested the workspace limits of the leg.
\item Object 3 - heavy and large - where the object was detectable but not movable due to its weight. This tested the reaction to immovable obstacles.
\end{itemize}

Object 1 and 2 were different shape and size, while Object 2 and 3 were the same shape and size but different weight. Figure~\ref{fig:all_objects} shows the different objects and how it appears to the perception system. Tests were predominately conducted on a flat carpet surface, with tests on marble and concrete to compare system performance on different surfaces.
Objects were placed the same distance away from Weaver for all the tests, but the robot had to adjust to the different object widths, detailed in Table~\ref{tab:object_details}.
Tests were mostly conducted indoors with consistent lighting conditions.

For all the tests, a single leg motion was used to move the object aside to clear the path in front of the robot. The motion consists of a quadratic and cubic B\'{e}zier curve executed in sequence. A quadratic curve was used to move the leg tip pose from the initialised pose to a predefined safe pose in front of the robot at an elevated height. Then a cubic curve was used to guide the leg tip to push the object aside, as shown in Figure~\ref{fig:high_vs_low_order_curve_1}. The positions for control points $P_1$ and $P_2$ in Figure~\ref{fig:weaver_control_point_test_sim} were influenced by the position of the object's key contact point.
 
The result of the leg motion was to move the obstacle in front of the robot away from it's path.

The Dynamixel motors are unable to provide a direct torque value, but have an effort value which is a ratio of the load experienced. An empirical estimate of this dimensionless effort output from the Dynamixel motors was used to determine when a leg was considered to be overloaded. For safety considerations, the limit was set below the estimate. If any of the joints for the selected manipulation leg is overloaded, the motion would cease its operation and return to a safe position so that the motors for the robot is protected from any damage. 

\subsection{Experimental Evaluation}
The performance for the system was based on the robot's ability to detect and move the object in front of it. A run was successful if the robot was able to detect and push aside the object, so that it no longer obscures the path for the robot to perform other actions, such as walking. The system was also successful if it was able to detect a potential motor overload and change its motion to prevent damage. A run was considered unsuccessful when the object was still in the robot's path  such that the robot would need to perform the motion again. It was also unsuccessful if the robot behaves in an undesirable way when interacting with the object, such as falling over. 

\section{Results}
\label{sec:results}

\begin{figure}[t]
     \centering
     \begin{subfigure}[b]{0.21\linewidth}
         \includegraphics[width=\linewidth]{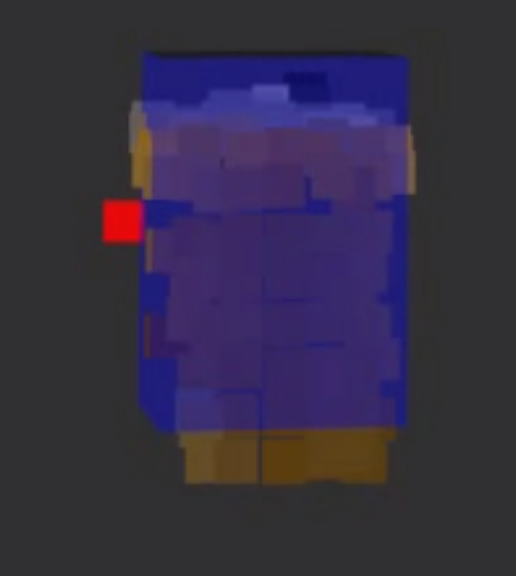}
         \caption{}
         \label{fig:rviz_pc_obj1}
     \end{subfigure}
     \hfill
     \begin{subfigure}[b]{0.34\linewidth}
         \includegraphics[width=\linewidth]{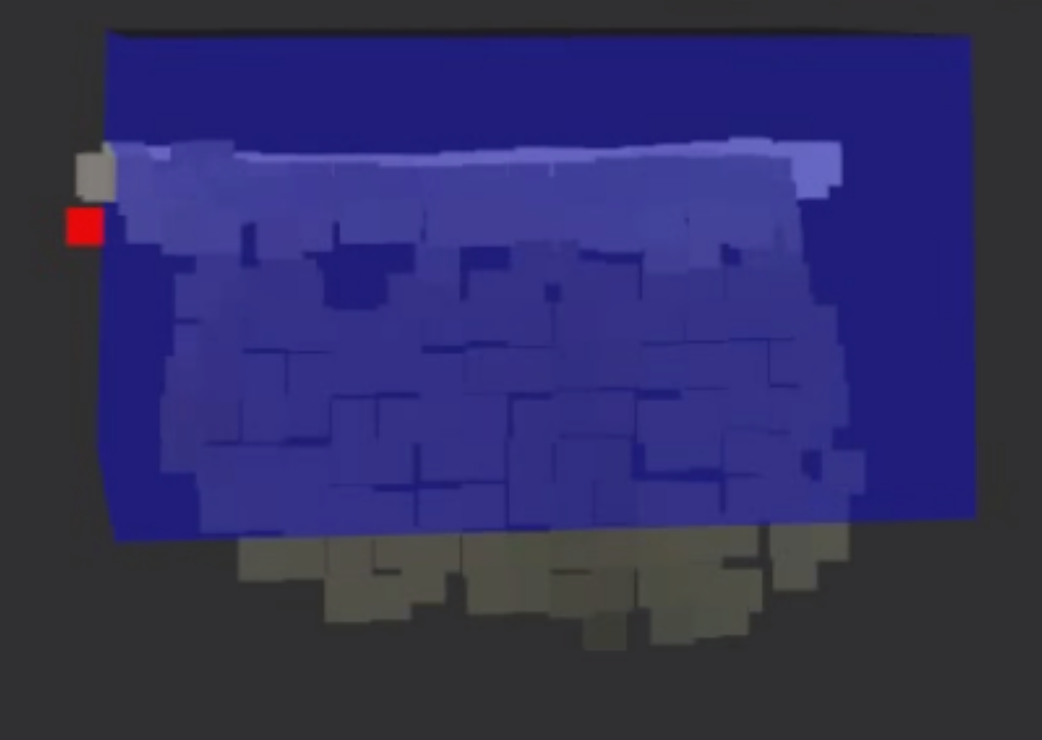}
         \caption{}
         \label{fig:rviz_pc_obj2}
     \end{subfigure}
     \hfill
     \begin{subfigure}[b]{0.35\linewidth}
         \includegraphics[width=\linewidth]{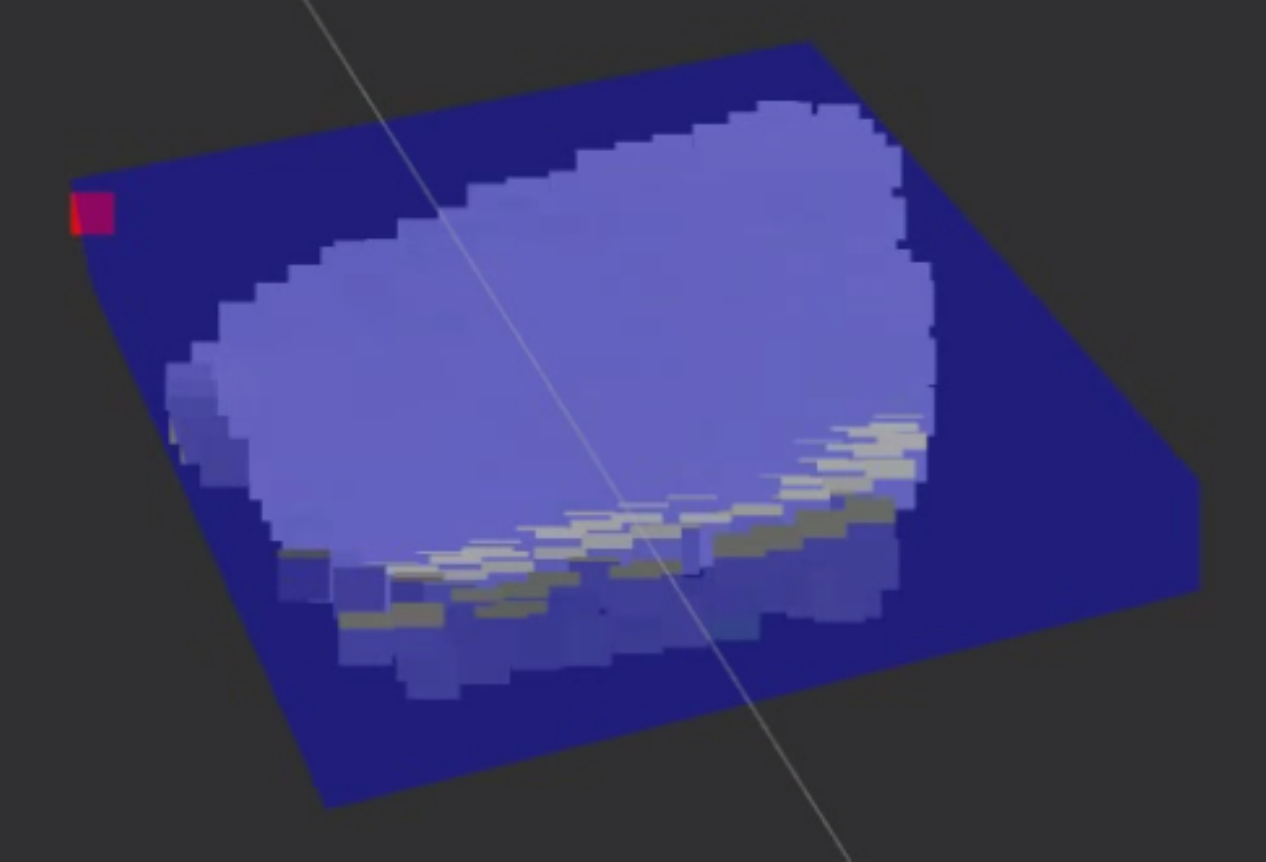}
         \caption{}
         \label{fig:rviz_pc_obj3}
     \end{subfigure}
        \caption{The key contact point (red dot) for Objects 1 (a), 2 (b) \& 3 (c), which influences the control points for the leg tip trajectory.}
        \label{fig:obj_points_pc}
\end{figure}

\begin{figure}[b!]
\centering
    \includegraphics[width=1\linewidth]{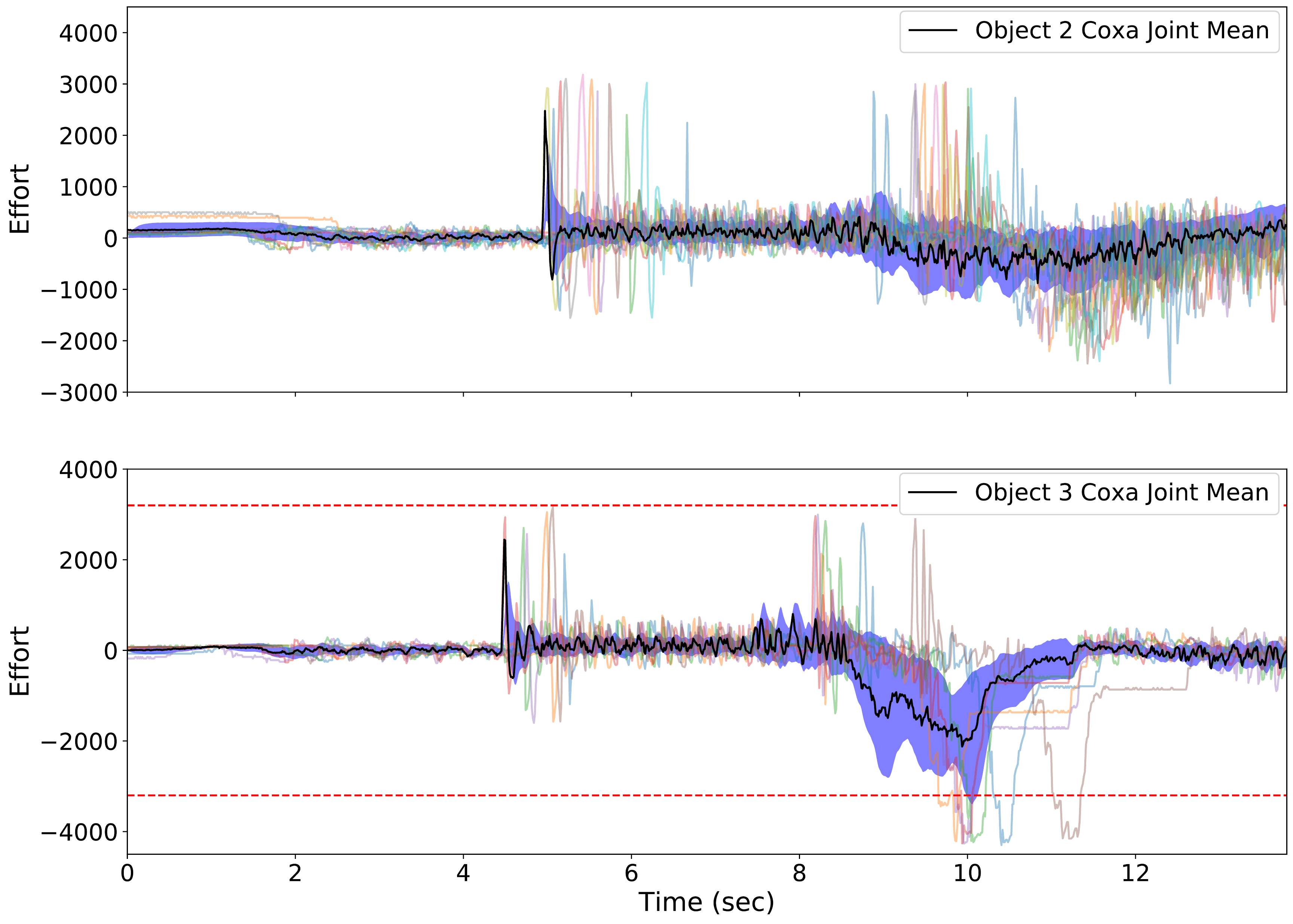}
    \caption{Effort exerted from coxa joint for Object 2 and 3. Positive and negative effort values indicates the direction of the joint. Effort limit threshold of -3200 is represented by the red dash line.}
    \label{fig:obj23_coxa_joints_plots}
\end{figure}

\begin{figure}[b!]
\centering
    \includegraphics[width=0.95\linewidth]{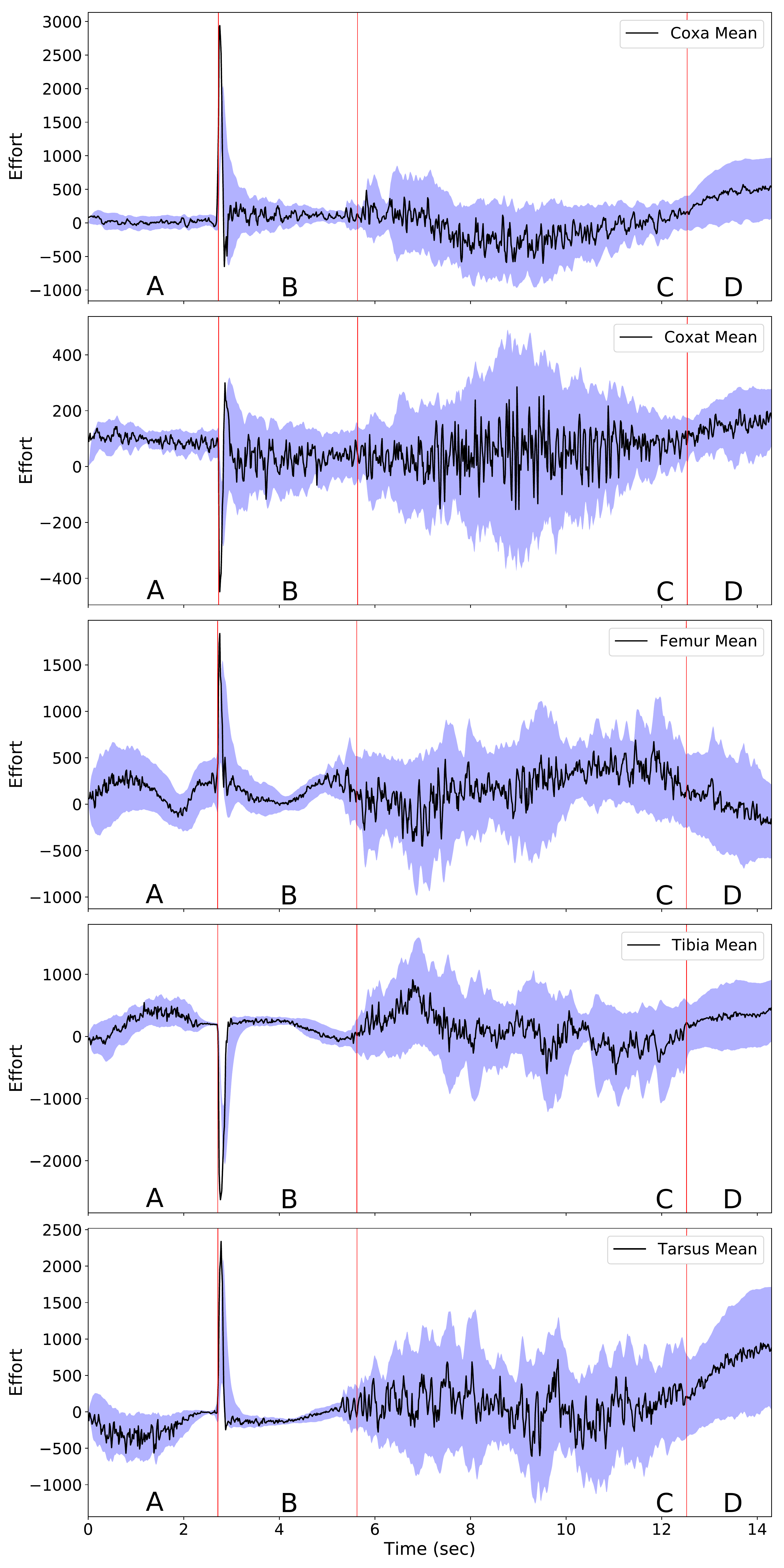}
    \caption {Mean and standard deviation of the manipulation leg's joint efforts for Object 1. A single standard deviation is represented by the purple shaded area. A: Leg initialising from stance to manipulation mode. B: Leg moving to in front of Weaver. C: Leg interacts with the object and pushes it aside. D: Leg returns to stance.}
    \label{fig:obj1_joints_plots}
\end{figure}

The perception system was successful in detecting all three objects, even though the environment was visually non-uniform as shown by the patches on the ground in Figure~\ref{fig:heroweaver}. The point cloud based approach was invariant to environment colour and lighting conditions, with observations of little disruption from lighting changes when the tests were run at different times of the day.
The system was able to provide object identification updates at over 20\,Hz.

The perception system provided the location of the desired key contact point. As shown by the red dot in Figure~\ref{fig:obj_points_pc}, this object contact point was located at point 5 based on the bounding box in Figure~\ref{fig:bounding_box}. The intermediate control points were influenced by the object's position so that the curve would pass through the object and that the behaviour of the curve was maintained as shown in Figure~\ref{fig:weaver_control_point_test_sim} for control points $P_1$ and $P_2$.

The generated leg trajectory points were both inside and outside the workspace. When the trajectory was inside the workspace, the desired position and orientation of the leg tip was achieved. In the case where the trajectory was beyond the reach of the robot's leg tip, emphasis was placed on following through with the behaviour of the trajectory, rather than achieving the desired leg tip pose.

For the heavy and large object scenario, the mass of the box was increased such that Weaver had difficulty moving it. Conducting experiments with the feedback from the robot's joint states, it was deduced that if any of the joints exceeded the effort value of $\pm$3500, the motors would enter protective shutdown. Thus, for the safety of the robot, the threshold was set to $\pm$3200. During execution, the motion would follow the trajectory as planned until the leg attempted to push the heavy box. Once the effort threshold was exceeded, the original leg motion was abandoned and a new set of trajectories was executed to return the leg back to stance position. Figure~\ref{fig:obj23_coxa_joints_plots} shows the comparison between the effort of the coxa joint during Object 2 and 3 scenario. The peak effort for every Object 3 trial exceeded the threshold, even though the mean did not exceed due to the alignment across the trials.

For both the light weight small and large object scenario, contact between the leg tip and the desired key contact point on the object was successful. Figure~\ref{fig:obj1_joints_plots} shows that most of the effort exerted is for maintaining the pose of the leg tip to remain normal to the ground and following the planned trajectory. Other surfaces listed in Table \ref{tab:object_details} were tested and yielded similar results as Object 2 when tested on carpet.

\section{Discussion}
\label{sec:discussion}
The perception system was designed so that the filters would remove any point cloud outliers that could interfere with the generation of the bounding box. Although the bounding box did not cover the entirety of the point cloud cluster, as shown in Figure~\ref{fig:obj_points_pc}, the data allowed the extraction of the object location and the key contact point for the leg tip.  

\begin{figure} [t]
\centering
    \includegraphics[width=1\linewidth]{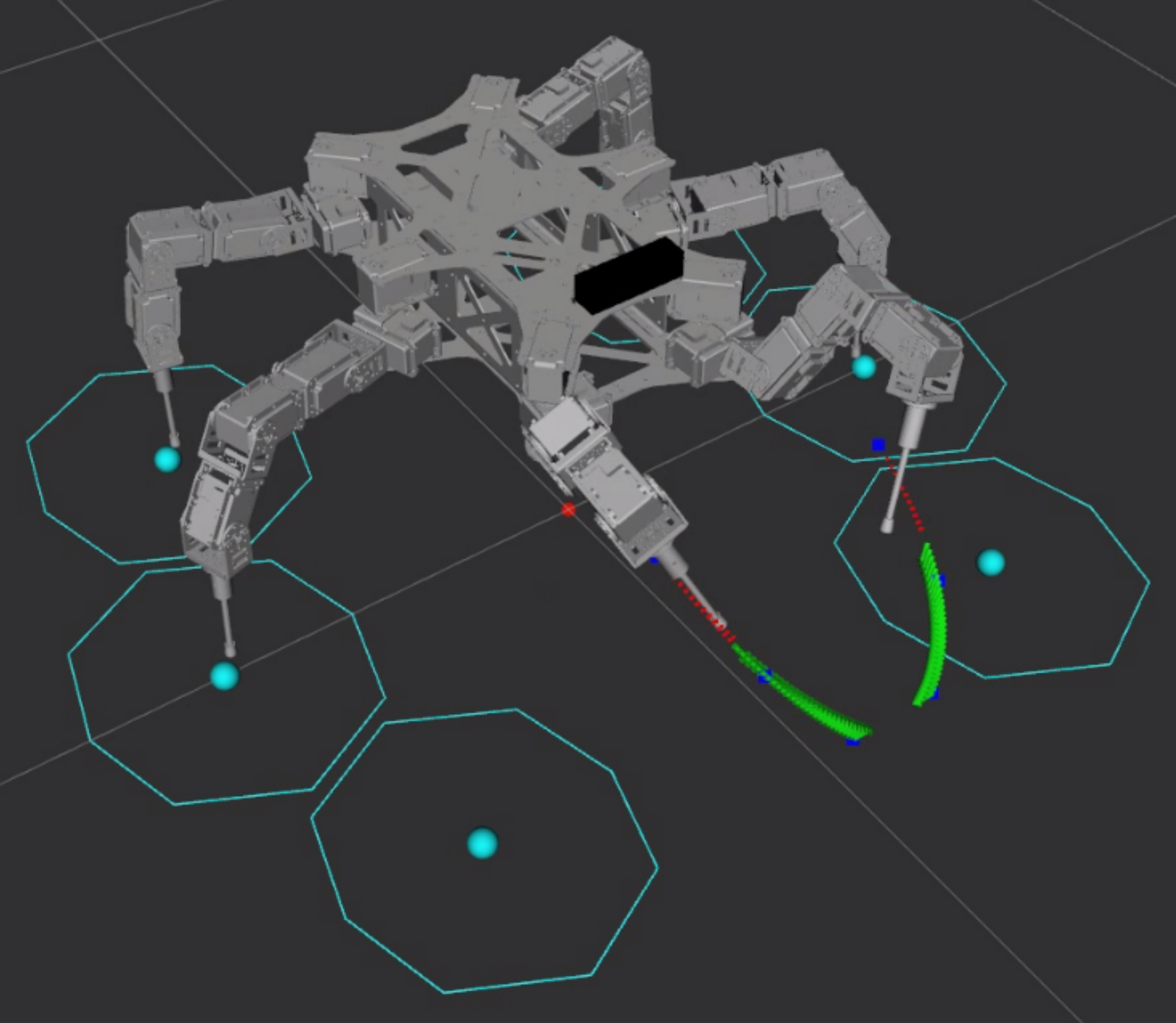}
    \caption{Unique trajectories simultaneously generated for both legs for advanced movements.}
    \label{fig:dual_leg}
            \vspace{-0.3cm}
\end{figure}

Object 2 and 3 compared the behaviour of the robot when faced with the same object shape and size but different weight. 

Object 3's weight was chosen to be movable but the system would experience instability. Heavier weights would result in the robot losing stability during manipulation and the coxa joint motor would overload and become nonfunctional for the rest of the movement, requiring a motor power cycle.

The leg overload detection prevented this occurrence as the system was able to sense the effort exerted by the manipulation leg and abort the motion when required. 
 
The type of movement what was used on all 3 scenarios was best suited for Object 2. For Object 2 and 3, the optimal point when being pushed aside was located in the centre depth of the object. Similarly, the same location relative to the object was used for Object 1. However, Object 1's reaction to this contact point was less favourable, resulting in the occasional rolling of the object. 

The option of moving two legs of the robot was explored, with the system able to create paths for simultaneous dual legipulation as shown in Figure~\ref{fig:dual_leg}. Each leg during dual legipulation had its own unique trajectory. However, this capability of more advanced legipulation behaviours such as grasping and lifting was only explored in simulation. Without onboard batteries, Weaver's altered centre of mass affected its stability. While not evident for the single leg tests, the large change in the support polygon for dual legipulation resulted in robot instability.

\begin{figure*}[ht!]
     \centering
         \begin{subfigure}[b]{0.48\linewidth}
         \includegraphics[width=\linewidth]{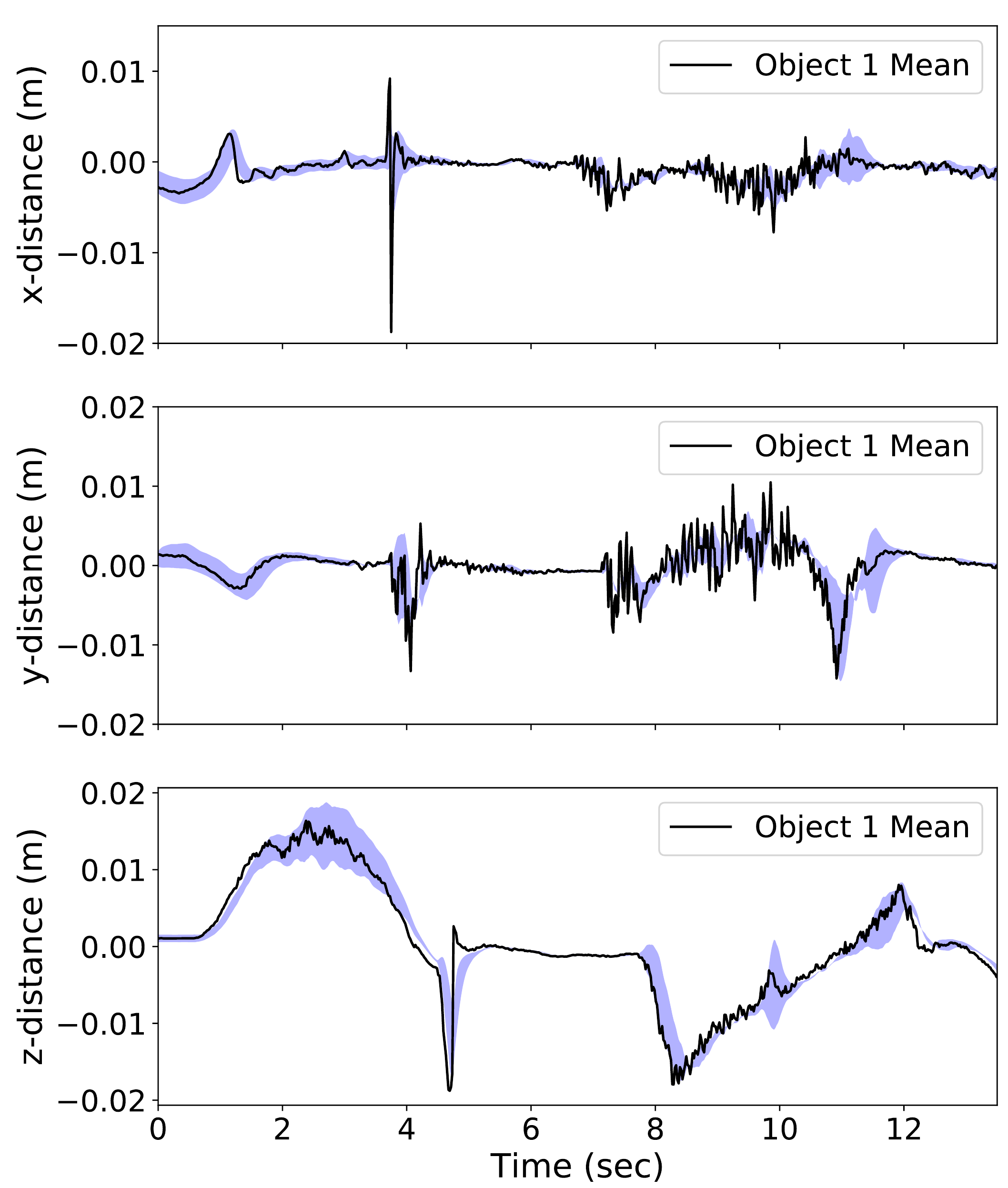}
         \caption{}
         \label{fig:diff_xyz_obj1}
     \end{subfigure}
     \begin{subfigure}[b]{0.48\linewidth}
         \includegraphics[width=\linewidth]{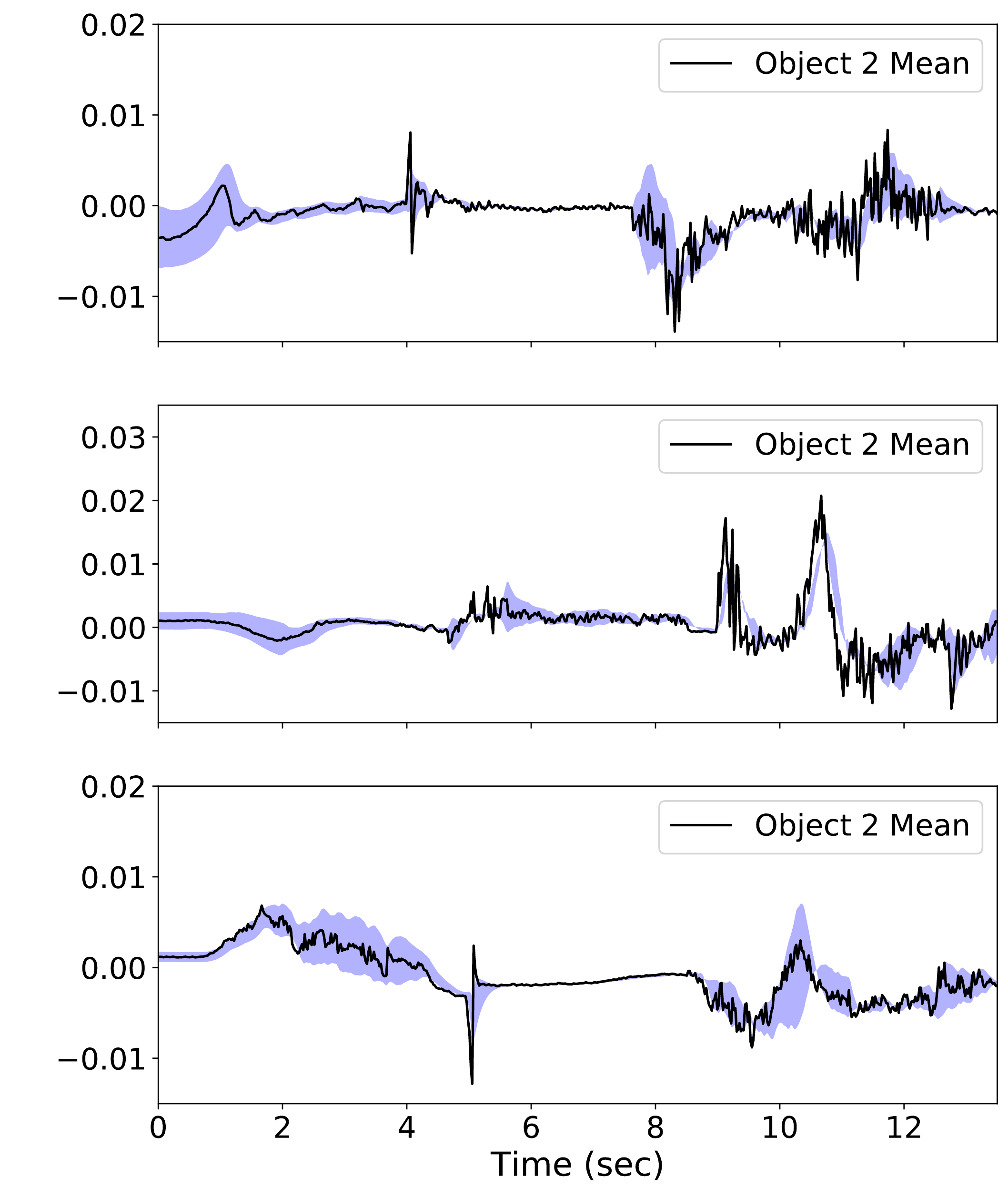}
         \caption{}
         \label{fig:diff_xyz_obj2}
     \end{subfigure}
        \caption{The mean of the difference between the desired and actual leg tip position for Object 1 (a) and 2 (b).}
        \label{fig:diff_xyz_obj}
        \vspace{-0.3cm}
\end{figure*}

The perception system was configured to view the area immediately in front of the robot, within the workspace of the legs. A visual servoing approach was not used due to the leg interfering with the camera's view during the leg motion sequence. Figure~\ref{fig:diff_xyz_obj} shows the difference in tracking between the desired and actual tip position, where the greatest error is less than 0.02\,m. The error was within the required accuracy for the intended purpose of the system, thus feedback was not required from the perception system during manipulation. That is, once the perception system provided the key contact point, legipulation was only controlled via proprioceptive feedback.

The behaviour of the robot was evaluated when interacting with objects on different ground surfaces. The friction between the ground and the object affects the performance of the system and would vary greatly, especially in disaster areas. Each surface tested had different levels of friction, with marble, concrete and carpet increasing respectively. Object 2 was tested on all these surfaces, with each surface yielding similar results. The robot was able to successfully move the object with little difficulty. 

\section{Conclusions}
\label{sec:conclusions}
This paper presented a method to control how legged robots manipulate objects within their environment. 
Key contact points for the obstacle extracted from the point cloud of the environment provided the information for the leg tip to successfully interact with the object. A combination of point cloud filters were used to create a bounding box around the obstacle for the key contact points, irrespective of the object's shape and size. With the ability to compose different leg sequences, unique movements can be created to best suit the situation. Experimental results show the system was able to generate a leg tip trajectory for a legged robot to follow, with a motion sequence that was influenced by the placement and shape of the object. Additionally, the leg overload detection safety module was successful in aborting the legipulation sequence when a heavy obstacle was present, preventing robot damage. 

In future works, we will consider the use of visual servoing to track the location the object during manipulation to adjust for any unexpected behaviours. Another goal includes exploring the potential of using any legs on the Syropod to manipulate objects, provided that the perception system covers the leg workspace.

\section*{Acknowledgments}
The authors would like to thank Fletcher Talbot for their support during the project. This work was fully funded by the Commonwealth Scientific and Industrial Research Organisation (CSIRO), Australia.

\balance
\bibliographystyle{named}

\end{document}